# Measurement of exceptional motion in VR video contents for VR sickness assessment using deep convolutional autoencoder

Hak Gu Kim
Image and Video System Lab.,
Electrical Engineering, KAIST,
South Korea,
hgkim0331@kaist.ac.kr

Wissam J. Baddar
Image and Video System Lab.,
Electrical Engineering, KAIST,
South Korea,
wisam.baddar@kaist.ac.kr

Heoun-taek Lim
Image and Video System Lab.,
Electrical Engineering, KAIST,
South Korea,
ingheoun@kaist.ac.kr

Hyunwook Jeong
Image and Video System Lab.,
Electrical Engineering, KAIST,
South Korea,
sceinstein@kaist.ac.kr

Yong Man Ro*
Image and Video System Lab.,
Electrical Engineering, KAIST,
South Korea,
ymro@kaist.ac.kr

## ABSTRACT

This paper proposes a new objective metric of exceptional motion in VR video contents for VR sickness assessment. In VR environment, VR sickness can be caused by several factors which are mismatched motion, field of view, motion parallax, viewing angle, etc. Similar to motion sickness, VR sickness can induce a lot of physical symptoms such as general discomfort, headache, stomach awareness, nausea, vomiting, fatigue, and disorientation. To address the viewing safety issues in virtual environment, it is of great importance to develop an objective VR sickness assessment method that predicts and analyses the degree of VR sickness induced by the VR content. The proposed method takes into account motion information that is one of the most important factors in determining the overall degree of VR sickness. In this paper, we detect the exceptional motion that is likely to induce VR sickness. Spatio-temporal features of the exceptional motion in the VR video content are encoded using a convolutional autoencoder. For objectively assessing the VR sickness, the level of exceptional motion in VR video content is measured by using the convolutional autoencoder as well. The effectiveness of the proposed method has been successfully evaluated by subjective assessment experiment using simulator sickness questionnaires (SSQ) in VR environment.

## CCS CONCEPTS

• **Human centered computing** → Virtual reality; • **General and reference** → Metrics; • **Machine learning** → Machine learning approaches

## KEYWORDS

Virtual Reality; Cybersickness; Machine Learning

## 1 INTRODUCTION

With the development of head mounted displays (HMDs), 360-degree camera, and 360 virtual reality (VR) content, interest in VR has been increasing for mass market applications including movies, games, training, etc [Bowman and McMahan 2007; Lécuyer et al. 2008]. VR content services can provide a unique viewing experience by immersing the viewer in a virtual environment. On the other hand, there has been an increasing concern about the safety issues of VR content [LaViola 2000], similar to 3D content [Jung et al. 2013a; Sohn et al. 2013; Jung et al. 2012; Jung et al. 2013b; Jeong et al. 2017]. Immersive VR video content could induce VR sickness, which is similar to motion sickness or cybersickness due to the discrepancy between the motion simulation and the viewer's motion (i.e., motion mismatches) [Hettinger and Riccio 1992; Von Mammen et al. 2016; Groen and Bos 2008]. Motion parallax [Baričević et al. 2012], field of view (FoV) [Lin et al. 2002], and time lag [Allison et al. 2001] are other reasons that can induce VR sickness as well. VR sickness can cause various symptoms: 1) oculomotor symptoms such as eye strain, blurred vision, and headache, 2) disorientation symptoms such as dizziness and vertigo, and 3) nausea symptoms such as salivation and stomach awareness [Kolasinski 1995; Carnegie and Rhee 2015]. These symptoms are similar to those of motion sickness [Arafat et al. 2016]. To proliferate VR content services, the viewing safety of the VR content is one of the most critical concerns. To address the problem of viewing safety in VR content services, it is essential to develop an objective VR sickness assessment (VRSA) method to predict the level of VR sickness of given VR video content. The objective VRSA can be utilized as safety guidelines in watching VR video contents by providing the degree of VR sickness of the content. In addition, it can be used to analyze VR video contents for the creation of VR video contents do not induce excessive VR sickness.

---
*\* Corresponding author*

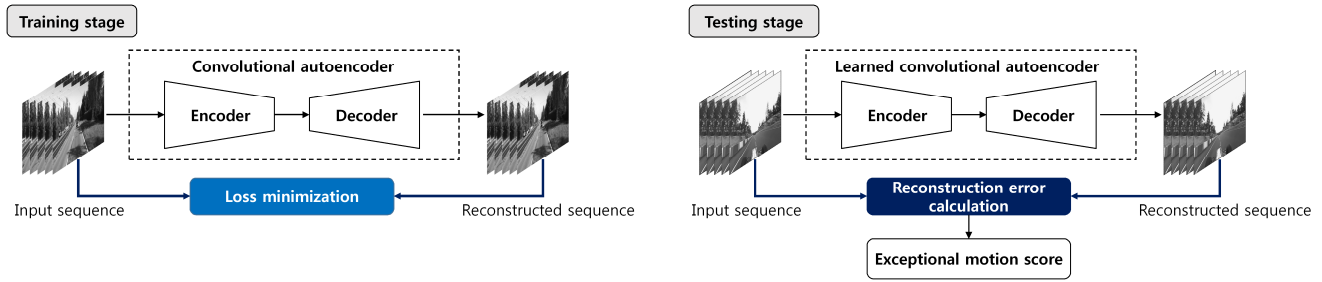

**Figure 1: Overview of the proposed method for measuring the effect of exceptional motion on the VR sickness using convolutional autoencoder network.**

There are many previous researches investigating causes of VR sickness (or cybersickness) in virtual environment. Most of the previous works focused on investigating the relationship between objective (e.g., physiological signals such as EEG, EGG, heart rate, etc.) and subjective assessment results (e.g., simulator sickness questionnaires (SSQ) [Kennedy et al. 1993]) whilewatching VR contents. To that end, meaningful VR sickness factors were studied [Kim et al. 2005; Dennison et al. 2016; Stauffert et al. 2016; Arthur 2000; Fernandes and Feiner 2016; Zielinski et al. 2015]. For example, it is well-known that the latency of display [Stauffert et al. 2016], speed of rendering [Arthur 2000], field of view (FoV) [Fernandes and Feiner 2016], and frame rate [Zielinski et al. 2015; Meehan et al. 2002] were some of the major factors affecting the VR sickness when watching VR contents with HMDs. Recently, Fernandes et al. showed that reducing the FoV could help to decrease VR sickness [Fernandes and Feiner 2016]. In [Carnegie, and Rhee 2015], experimental results demonstrated that depth of field (DoF) was one of main causes related to the overall visual discomfort of the VR content by comparing SSQ scores with DoF blur enabled and disabled. In [Dennison et al. 2016; Zielinski et al. 2015; Meehan et al. 2002], the effects of frame rate in virtual environment were explored. Many user studies with SSQ scores reported that low frame rate could negatively impact VR sickness because motion continuity could be lost in low frame rate. In [Egan et al. 2016], a subjective and objective Quality of Experience (QoE) evaluation was presented for immersive VR content displayed through HMDs. However, most of previous VR sickness studies focused on investigating and analyzing the relation between objective physiological signals and subjective assessment results.

In this paper, a novel deep learning-based objective measurement framework is proposed to analyze the exceptional motion (i.e., fast motion velocity) of VR content for VR sickness assessment. The exceptional motion in VR video content is one of the most important factors inducing the VR sickness because it could exacerbate motion mismatches between simulation motion and viewer's motion [Hettinger et al. 1992; Von Mammen et al. 2016; Groen and Bos 2008; Riva et al. 1998]. We devise a deep convolutional autoencoder to measure the exceptional motion. The convolutional autoencoder is trained to reconstruct original VR video sequences (video datasets with non-exceptional motion such as slow and moderate motion velocity). Thus, the trained autoencoder model reconstructs VR video sequence with non-exceptional motion well. On the other hand, VR video sequence with exceptional motion, which is reconstructed by the trained autoencoder model, shows large reconstruction errors. In the proposed method, by measuring the reconstruction error of the motion information in VR video content, the exceptional motion of VR video contents can be detected and measured. Based on the measured exceptional motion by our deep network, the level of VR sickness of VR video content can be predicted. Hence, with the proposed objective metric can reduce the labor-intensive effort and lots of time for objectively assessing the VR sickness. Experimental results show that the proposed measurement is highly correlated with the human subjective scores.

The rest of this paper is organized as follows. In Section 2, the proposed method for VRSA is described. In Section 3, experiments and results are shown to demonstrate the performance of the proposed method. Finally, conclusions are drawn in Section 4.

## 2 PROPOSED METHOD

Figure 1 shows an overview of the proposed method for objectively assessing the VR sickness of VR video contents. As shown in the figure, the proposed method is based on deep convolutional autoencoder network. The network consists of encoder for feature extraction and decoder for reconstruction from encoded features. In the training stage, the convolutional autoencoder model is learned to reconstruct input consecutive frames by minimizing the loss between original input frames and reconstructed output frames. Because the deep network is trained with a large scale of video dataset with small motion or moderate motion, the trained model can reconstruct the original VR video sequences with non-exceptional motion but cannot reconstruct the VR video content with exceptional motion in the testing stage. As a result, by measuring the exceptional motion score based on reconstruction quality of the proposed network, the degree of VR sickness of VR video content can be predicted without additional information such as physiological measurements. Detailed descriptions are given in the following sections.

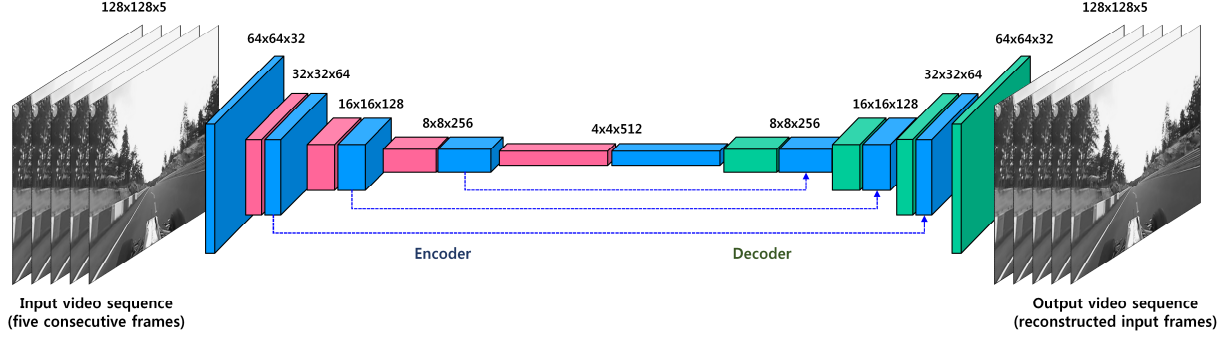

**Figure 2: Proposed convolutional autoencoder network for measuring the exceptional motion of VR video content. Blue and green layers indicate convolution layers and deconvolution layers, respectively. Red layers represent max pooling.**

## 2.1 Deep Convolutional Autoencoder for Normal Motion Patterns Learning

As abovementioned, there are various factors inducing VR sickness [Kim et al. 2005; Dennison et al. 2016; Stauffert et al. 2016; Arthur 2000; Fernandes and Feiner 2016; Zielinski et al. 2015; Meehan et al. 2002; Egan et al. 2016; Riva et al. 1998]. In particular, exceptional motion can lead to severe VR sickness while watching the VR video contents since it exacerbates motion mismatches between perceived locomotion and physical steadiness.

In this paper, a deep convolutional autoencoder network is employed to detect and measure the exceptional motion patterns. Figure 2 shows the proposed convolutional autoencoder network, aiming to reconstruct the VR video contents with non-exceptional motion. To that end, the proposed network consists of two main parts which are convolutional encoder and decoder, as shown in the Fig. 2. Five consecutive frames are used as input data. The convolutional encoder is to encode multi-level deep features from input frames. For this purpose, this part comprises 5 convolution layers and 5 max pooling layers. In the encoder part, all filters (size of 3 x 3) are learned in convolution layers. The maximum pool layers with size of 2 x 2 with stride 2 are employed to reduce the spatial resolution of the feature map at each layer by half. As the spatial resolution of the feature map is reduced by half, the number of channels of the feature map is doubled. By convolution and max pooling operations, multi-level features of the input frames (5 consecutive frames) are encoded. The encoded features from encoder part are fed to the decoder part in order to reconstruct the input VR video sequence. The role of convolutional decoder is to generate the input sequence from encoded deep features at convolutional encoder. In our deep network, the decoder part consists of deconvolution layers. The deconvolution layers help to generate the target VR video sequence by decoding the deep features. In the decoder part, all filters with size of 3 x 3 at each layer are learned for deconvolution operation. In addition, a skip connection is applied to our network for high quality reconstruction. It can mitigate the effect of loss of the detail information on the spatial domain by max-pooling operation. In Fig. 2, the blue dotted arrows indicate skip connection process. It links between convolution and deconvolution layers by concatenating feature maps of convolution layers and those of deconvolution layers. As shown in Fig. 2, in the decoder part, the resolution of feature map is doubled, as the number of channels of the feature map is reduced by half. Finally, we obtain the reconstructed five consecutive frames with the same size of the input.

In the training stage, the convolutional auto encoder is trained by minimizing the errors between input frames and reconstructed frames. For training, Euclidean loss between input frames and reconstructed frames is calculated.

$$\hat{f}_W = \underset{W}{\operatorname{argmin}} \frac{1}{2N} \sum_{k=1}^{N} \sum_{i=1}^{W} \sum_{j=1}^{H} \| X(i,j,t) - f_W(X(i,j,t)) \|^2 \quad (1)$$

where $W$ and $H$ are width and height of training images, respectively. $N$ is the mini batch size. $X(i, j, t)$ is $t$-th frame. $i$ and $j$ are pixel indices, respectively. $f_W(\cdot)$ is the non-linear function, which is a convolutional autoencoder network with weights $W$.

## 2.2 Exceptional Motion Pattern Score

After training our convolutional autoencoder network, the exceptional motion pattern score can be calculated based on the reconstruction errors, similar to [Hasan et al. 2016]. With the learned model, the reconstruction error at $t$-th frame of the test datasets is denoted as $e(t)$, which can be written as

$$e(t) = \sum_{i=1}^{W} \sum_{j=1}^{H} \| I(i,j,t) - \hat{f}_W(I(i,j,t)) \|^2 \quad (2)$$

where $\hat{f}_W$ is the learned function by our convolutional autoencoder model. $I(i, j, t)$ is $t$-th frame of test datasets. Let $s_m(t)$ denote an exceptional motion pattern score. Then, the final exceptional motion pattern score at $t$-th frame of the test image, $s_m(t)$, is given by

$$s_m(t) = \frac{e(t)}{\sqrt{W \times H}} \quad (3)$$

Lower exceptional motion pattern score indicates that the frame is not likely to induce VR sickness because it can be reconstructed well by the trained deep network. On the other hand, a higher exceptional motion pattern score indicates that the frame is highly likely to lead to VR sickness.

VRST '17, Gothenburg, Sweden    Hak Gu Kim et al.

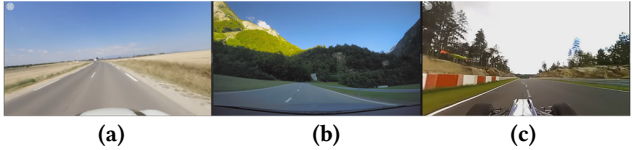

(a)                    (b)                    (c)

**Figure 3: 360-degree VR videos used in our experiments. (a) VR video1 with slow motion, (b) VR video2 with moderate motion, and (c) VR video3 with fast motion.**

## 3 EXPERIMENTS AND RESULTS

### 3.1 Datasets for Learning the Deep Network

In our experiment, to train the convolutional autoencoder, we used various video datasets, which were UCSD Ped1 and Ped2 [Mahadevan et al. 2010], Avenue datasets [Lu and Jia 2013], and KITTI benchmark datasets [Geiger et al. 2013]. In UCSD Ped1, there are 34 training video clips. Each video clip contains 200 frames. In UCSD Ped2, there are 16 training video clips. The number of frames at each video clip is different. The number of frames ranges between 120 and 180 frames. Avenue dataset has 16 training video sequences. Each video sequence duration varies between less than one minute to two minutes. In KITTI benchmark, city, road, and residential driving datasets were used as training data. A total 61 video clips were used for training (61 video clips = 28 city videos+21 residential videos+12 road videos). The resolution of KITTI benchmark dataset is 1242 x 375 pixels (rectified version). The number of frames is very different according to each clip. Its details refer to [Geiger et al. 2013].

In our experiment, all images of UCSD Ped1, 2, and Avenue were resized to 128 x 128 for memory efficiency and an effective computational load. These three video datasets were captured by fixed camera environment. They were used for pre-training of our model. Based on the pre-trained model, our convolutional autoencoder was fine-tuned by training the model using a large scale of KITTI benchmark datasets. All images of KITTI benchmark were resized to 50% and randomly cropped with a size of 128 x 128 pixels. In our experiment, the training datasets were temporally augmented to increase the size of the training data. To augment the training data in temporal direction, we generated the input video frames with three different temporal stride 1, 2, and 3. For example, the generated videos with stride 2 have following frame numbers [1, 3, 5, 7, …] and [2, 4, 6, 8, …].

### 3.2 Datasets for Test

To evaluate the performance of the proposed method and measure the VR sickness in subjective assessment, we used publicly available three 360-degree VR video contents, collected from Youtube[1,2,3]. Fig. 3 shows the captured scenes in each 360-degree VR video content used in our subjective assessment experiment. As shown in Fig. 3, they were captured during driving on a road. They had different motion patterns for three different scenarios. Average motion in the first video (Fig. 3 (a)) was slow and the second VR video content had moderate motion velocity (Fig. 3 (b)). In the third VR video content captured in racing car (Fig. 3 (c)), its average motion was very fast.

**Table 1: SSQ used in our subjective assessment**

| SSQ Symptoms | Nausea | Oculomotor | Dis-orientation |
|---|---|---|---|
| General discomfort | O | O | |
| Fatigue | | O | |
| Headache | | O | |
| Eye strain | | O | |
| Difficulty focusing | | O | |
| Increased salivation | O | | O |
| Sweating | O | | |
| Nausea | O | | |
| Difficulty concentrating | O | O | O |
| Fullness of head | | | |
| Blurred vision | | O | O |
| Dizzy (Eyes open) | | | O |
| Dizzy (Eye closed) | | | O |
| Vertigo | | | O |
| Stomach awareness | O | | O |
| Burping | O | | |

### 3.3 Experimental Setup

Our experiments for training and testing were conducted on a PC with Intel Core i7-4770 @ 3.4 GHz, 32GB RAM, and NVIDIA GTX 1080. The proposed convolutional autoencoder model for exceptional motion analysis was implemented by TensorFlow. For network learning, 5-fold cross validation was used. For each validation set, the datasets were randomly divided into 80% for training and 20% for testing.

### 3.4 Subjective Assessment Experiments

*3.4.1 Equipment for Displaying VR Contents.* In our subjective assessment experiment, we used Oculus Rift CV1 HMD with Intel Core i7-4770@3.4 GHz, 32GB RAM, and NVIDIA GTX 1080TI in order to present 360 degree VR video content. The Oculus Rift CV1 presented the VR video contents with a spatial resolution of 2160 x 2400 pixels (1080 x 1200 pixels per each eye). Its frame rate is 90Hz and FoV is about 100 degree.

*3.4.2 Subjects for Subjective Assessment Experiments.* A total 17 subjects, ranging between 20 to 30 years of age, participated in our subjective assessment experiments. Two subjects were excluded because they were detected as outliers by screening methodology recommended in [ITU-R 2002]. As a result, with a total of 15 subjects, our experiment was conducted. They had normal or corrected-to-normal vision and had a minimum stereopsis of 60 arcsec. The average stereopsis of the subjects was 37.50±4.25 arcsec, which was assessed with the Randot stereo                                                                    test.

---

[1] https://www.youtube.com/watch?v=JEr3-FzSgzk
[2] https://www.youtube.com/watch?v=wECZs7hewjY
[3] https://www.youtube.com/watch?v=wfNvZwN87Hg

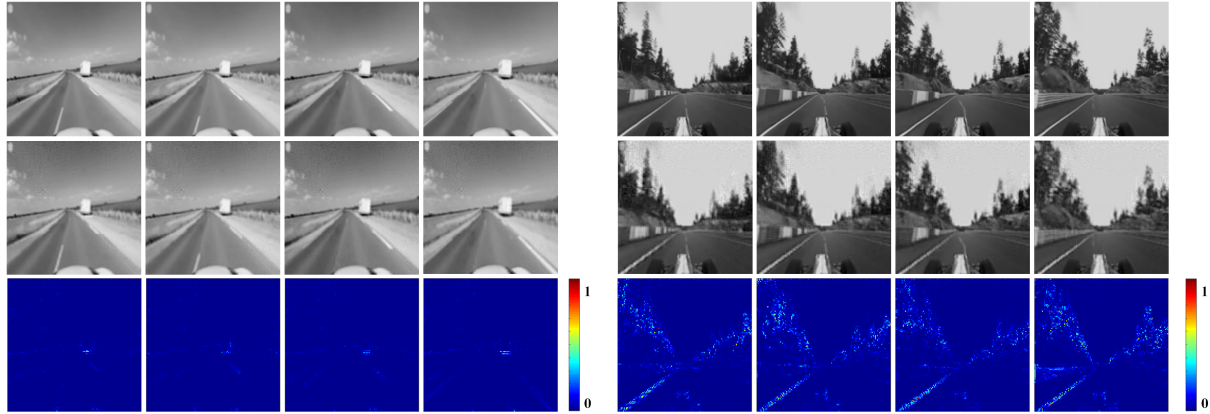

**Figure 4: Reconstruction results of the proposed deep convolutional autoencoder network for VR video 1 with slow motion (left) and VR video 3 with exceptional motion (right). First and second rows indicate original input frames and reconstructed frames by the deep network, respectively. The third row indicates reconstruction error maps between original frames and reconstructed frames. Note that all images are normalized in range of [0, 1].**

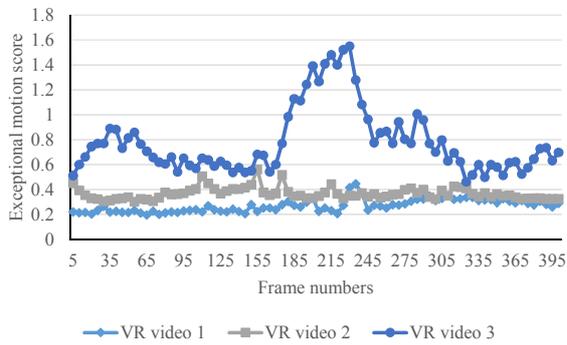

**Figure 5: Exceptional motion score for each VR video.**

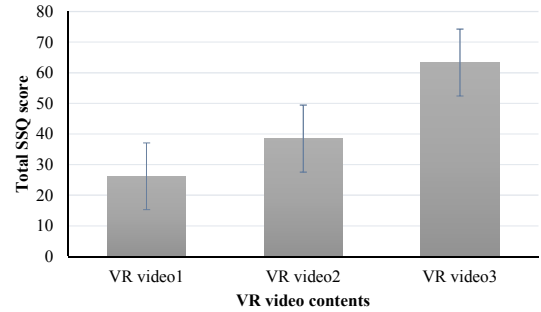

**Figure 6: Measured total SSQ score for each VR video.**

*3.4.3 Questionnaires.* In our subjective assessment experiments, 16-item SSQ (latest version) was used to measure the degree of VR sickness in watching the VR video contents, as shown in Table 1 [Kim et al. 2005; Rebenitsch and Owen 2016]. For the assessment of VR sickness, we asked participated subjects to complete the SSQ sheet before and after being exposed to three VR video contents [Chessa et al. 2016]. The subjects used a discrete scale divided into four levels in order to grade the VR sickness. The labels of SSQ were 'None', 'Slight', 'Moderate', and 'Severe'. For each symptom, the scores were 0 for 'None', 1 for 'Slight', 2 for 'Moderate', and 3 for 'Severe' [Kim et al. 2005; Chessa et al. 2016]. Finally, a total SSQ score was calculated by combining every partial scores for each symptom with the weight, which was set to 3.74 [Chessa et al. 2016]. Note that the total SSQ score ranging of 32 to 40 is enough to indicate perceptible cybersickness [Kim et al. 2005; Bruck and Watters 2009].

*3.4.4 Procedures.* To evaluate the VR sickness in VR viewing with HMD, we conducted subjective assessment experiments with three VR video contents including different motion patterns. A week before the actual subjective assessment experiments, we had subjects experience a variety of VR contents with Oculus Rift in order to allow them familiarize with VR environment.

Every VR video content was displayed for 2 minutes through Oculus Rift CV1. After watching the VR video contents, subjects rated their perception of the VR sickness for each symptom in SSQ sheet instructed at the beginning of the experiment. Each subjective assessment experiment for each VR video content was conducted on a different day for viewing safety of the participating subjects. During the subjective assessment, the subjects could immediately stop and take a break if they felt any difficulty to continue due to excessive VR sickness. The test was conducted under approval from Institutional Review Board (IRB).

### 3.5 Performance Evaluation Results

*3.5.1 Performance of the proposed deep convolutional autoencoder for exceptional motion measurement.* To evaluate the performance of the proposed deep network, we performed the reconstruction of three VR video contents as shown in Fig. 3. Figure 4 shows the reconstruction results of the proposed convolutional autoencoder network for VR video 1 and 3. In Fig. 4, the first and second rows indicate the original input frames and output frames reconstructed by the proposed deep network. The last row indicates the reconstruction error maps between the original and reconstructed frames. As shown in the Fig. 4, VR video 1 with slow motion (left of Fig. 4) was reconstructed well. On the other hand, the reconstruction quality of VR video 3 was worse than that of the VR video 1 because it was more difficult to reconstruct exceptional motion than slow motion patterns. Figure5 shows a plot of proposed exceptional motion score for the three VR videos. Exceptional motion score of VR video 1 was the lowest while VR video 3 had the highest exceptional motion score at a number of frames.

*3.5.2 Subjective Assessment Results.* To evaluate the level of VR sickness of each VR video content, we conducted subjective assessment using SSQ, described in Section 3.4. Figure 6 shows the results of our subjective assessment for three VR video contents. As shown in the Fig. 6, the degree of VR sickness subjects felt was proportional to the motion magnitude. In general, previous studies reported that total SSQ score of range 32 to 40 could be enough to cause cybersickness [Kim et al. 2005; Bruck and Watter 2009]. Hence, the results of subjective assessment showed that subjects felt some symptoms of VR sickness when watching VR video 2 and 3. In particular, VR video 3 could lead to excessive VR sickness.

*3.5.3 Performance evaluation of the proposed method for assessing the VR sickness.* Finally, in order to verify the performance of the proposed exceptional motion measurement for objectively assessing the VR sickness, Pearson linear correlation coefficient (PLCC) was employed. It was one of the commonly used performance measures. The PLCC between our exceptional motion score and total SSQ score was 0.92. Thus, the evaluation results revealed that the proposed method was highly correlated with subjective SSQ scores of our VR video contents. Importantly, these results indicated that the exceptional motion is one of the most important factors on the VR sickness of VR video content.

## 4 CONCLUSIONS

This paper presented a novel measurement of exceptional motion using deep convolutional autoencoder network for assessing the VR sickness of VR video content. The convolutional autoencoder learned by normal datasets with slow and moderate motion could reconstruct the non-exceptional motion patterns but it could not recover VR video contents having exceptional motion. Based on the fact that exceptional motion led to high reconstruction errors in the deep autoencoder network, the level of VR sickness of the input VR video content due to exceptional motion could be predicted. The results of our subjective assessment experiments showed that the proposed objective measure strongly had a high correlation with human subjective quality scores, SSQ of our test datasets (PLCC was 0.92). This result indicated that exceptional motion velocity was one of the critical factors of the VR sickness.